\documentclass[journal]{IEEEtran}
\usepackage{times}

\usepackage{soul}
\usepackage{url}
\usepackage[draft]{hyperref}
\usepackage[utf8]{inputenc}
\usepackage[small]{caption}
\usepackage{graphicx}
\usepackage{amsmath}
\usepackage{booktabs}
\usepackage{bm}
\usepackage{amsmath}
\usepackage{booktabs}
\usepackage{amssymb}
\usepackage{longtable}
\usepackage{multirow}
\usepackage{subfigure}
\usepackage{array}
\usepackage{color}

\ifCLASSINFOpdf
\else
\fi
\begin{document}
%
\title{Action Shuffling for Weakly Supervised Temporal Localization}
%
%

\author{Xiao-Yu Zhang,~\IEEEmembership{Senior Member,~IEEE,}
        Haichao Shi,
        Changsheng Li,~\IEEEmembership{Member,~IEEE,}\\
        and~Xinchu Shi
\thanks{X.-Y. Zhang and H. Shi are with Institute of Information Engineering, Chinese Academy of Sciences, Beijing, China, 100093. H. Shi is also with School of Cyber Security, University of Chinese Academy of Sciences, Beijing, China. (e-mail: zhangxiaoyu@iie.ac.cn; shihaichao@iie.ac.cn)}
\thanks{C. Li is with Beijing Institute of Technology, Beijing, China. (e-mail: changshengli507@163.com)}
\thanks{X. Shi is with Meituan Group, Beijing, China. (e-mail: shixinchu@meituan.com)}
}

%
%

\markboth{Journal of \LaTeX\ Class Files,~Vol.~14, No.~8, August~2015}%
{Shell \MakeLowercase{\textit{et al.}}: Bare Demo of IEEEtran.cls for IEEE Journals}
%



\maketitle

\begin{abstract}
Weakly supervised action localization is a challenging task with extensive applications, which aims to identify actions and the corresponding temporal intervals with only video-level annotations available. This paper analyzes the order-sensitive and location-insensitive properties of actions, and embodies them into a self-augmented learning framework to improve the weakly supervised action localization performance. To be specific, we propose a novel two-branch network architecture with intra/inter-action shuffling, referred to as ActShufNet. The intra-action shuffling branch lays out a self-supervised order prediction task to augment the video representation with inner-video relevance, whereas the inter-action shuffling branch imposes a reorganizing strategy on the existing action contents to augment the training set without resorting to any external resources. Furthermore, the global-local adversarial training is presented to enhance the model’s robustness to irrelevant noises. Extensive experiments are conducted on three benchmark datasets, and the results clearly demonstrate the efficacy of the proposed method.
\end{abstract}


%
\IEEEpeerreviewmaketitle

\section{Introduction}
\IEEEPARstart{T}{emporal} action localization is one of the most challenging tasks in video content understanding, which has attracted intensive attention in the community. Given an untrimmed video, action localization aims to identify the time intervals corresponding to the actions of interest. Remarkable progress has been made in the fully supervised scenario~\cite{ssn,chao,BSN,zeng2019graph}, where frame-level annotations are indispensable. Unfortunately, the overwhelming labeling effort to obtain the detailed annotations renders the fully supervised methods inapplicable to large-scale video sets. This leads to the prevalence of weakly supervised paradigm~\cite{w-talc,CMCS,nguyen2019weakly,pretrimnet,DGAM,BasNet,tscn}, which only requires video-level annotations to deduce frame-level predictions. To date, various weakly supervised action localization methods have been put forward and the most recent advances manifest two striking trends. (1) \textbf{Action-background modeling}. As indicated by the latest works, explicitly modeling the action and background contents proves to be an effective way of representation learning~\cite{CMCS,nguyen2019weakly,DGAM,pretrimnet}. By learning the video-level action and background representations of a video separately, action localization performance can be improved collaboratively. However, the video-level modeling can only capture coarse-grained description. Less study has been directed on in-depth analysis of the intrinsic properties of actions. (2) \textbf{Exploration of external resources}. In order to make up for the limited information in weak supervision, it has become another trend to resort to external resources. Typically, the publicly available videos or generated pseudo videos with video-level or frame-level labels are leveraged as supplementary training data~\cite{TSRNet,CMCS,actionbytes}. For all the performance gain achieved by this manner, new challenges have also emerged. For one thing, source-target adaptation between the original and auxiliary dataset is vital but difficult for robust knowledge transfer. For another, feature extraction of the new training videos lays extra burden on computational consumption. To address the above issues, this paper aims to reveal the properties of actions and embody them into the model to achieve effective and efficient weakly supervised action localization.

\begin{figure}[tbp]
\centering
\includegraphics[width=1\linewidth]{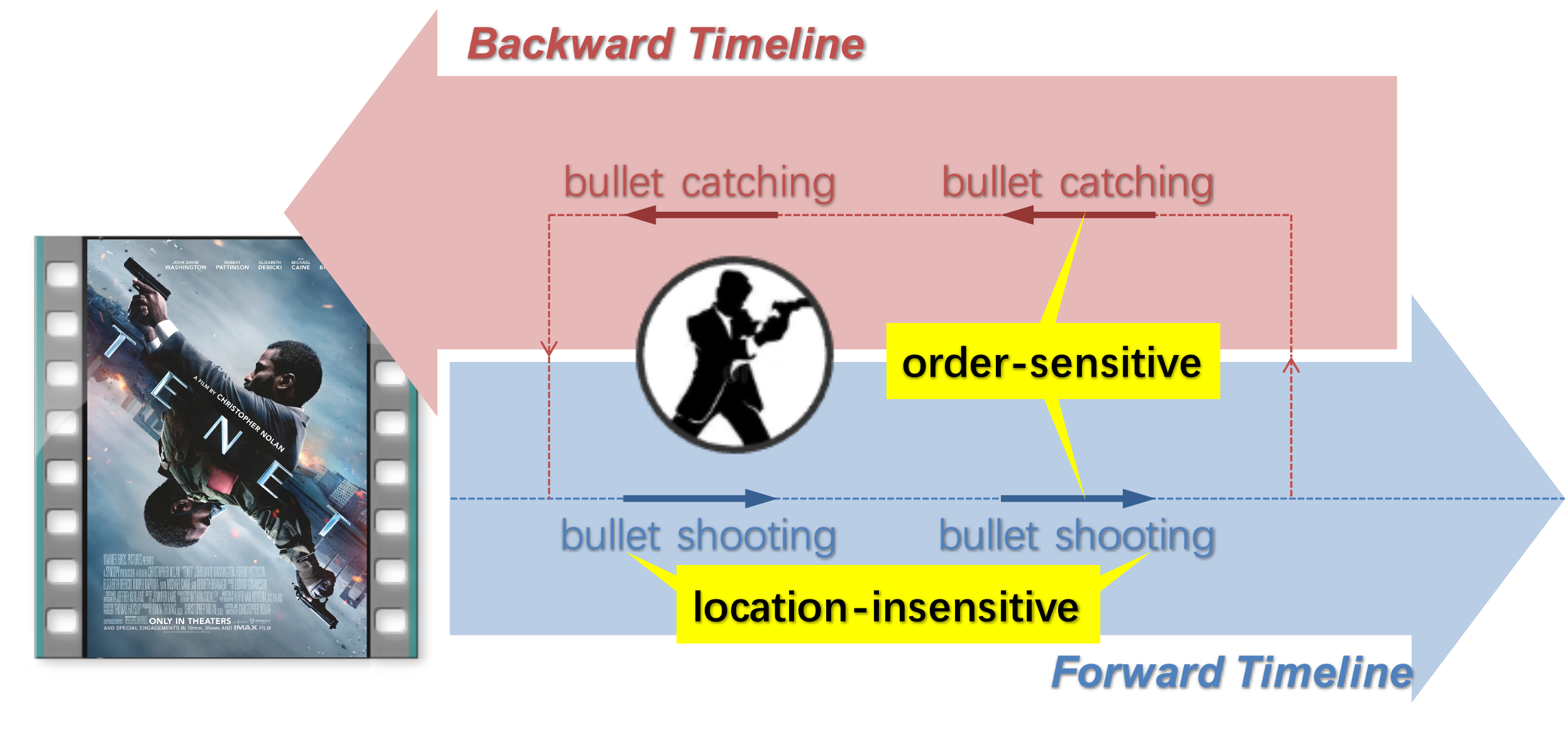}
\caption{Illustration of the order-sensitive and location-insensitive properties of actions.}
\label{fig1}
\end{figure}

Tenet, a 2020 hit movie directed by Christopher Nolan, opens up a dramatic world with the art of ``time inversion", where the characters can go back in time. This paper does not intend to be a spoiler of the story. However, there are two interesting phenomena which enlighten us greatly in action analysis. In the movie, when going backward in time, actions become weird and hard to understand. In contrast, when a time point in the past has been reached via backward time travelling and the forward timeline is restored, actions get back to normal and seem to blend in well although the surrounding environments have changed drastically. This mind-blowing movie indicates two critical properties of actions illustrated in Fig.~\ref{fig1}.

\begin{itemize}
  \item On one hand, actions are \textit{order-sensitive}. As we know, the dynamic motional properties of a video are reflected by the temporal relevance within ordered frames. Altering the inner order of an action may significantly change its semantics. Particularly, in Tenet, bullet shooting in reversed order becomes bullet catching, which is an utterly different action.
  \item On the other hand, actions are \textit{location-insensitive}. Compared to the dependence on inner order, an action is relatively independent on when it takes place. Taking actions of the same category in different time points is not likely to affect the underlying semantics, as long as the original inner order is retained.
\end{itemize}

Inspired by the order-sensitive and location-insensitive properties of actions, in this paper, we propose a novel weakly supervised action localization network architecture with intra/inter-action shuffling, referred to as ActShufNet. On top of the conventional attention-based action recognition and localization paradigm, we build a self-augmented learning model to achieve improved representative ability, without resorting to any external resources. Starting with the preliminarily segmented actions based on class-agnostic attentions, our model goes through two pipelines, i.e. the intra-action and inter-action shuffling. Intra-action shuffling randomly alters the inner order of an action and aims to restore its original order via a self-supervised task. In this way, the optimized representations are forced to capture the underlying inner relevance of actions, which facilitate the subsequent semantic deduction. Inter-action shuffling randomly picks actions of the same category and collectively creates new untrimmed videos which are naturally attached with the shared video-level labels. In this way, the training dataset can be arbitrarily expanded, and meanwhile more variety is included within each created video. To further enhance the model’s discriminative ability between action and background, global-local adversarial training scheme is presented to achieve perturbation tolerable robust learning performance. The main contributions of our work are summarized as follows.

\begin{itemize}
  \item We develop the intra/inter-action shuffling mechanism to fully exploit the order-sensitive and location-insensitive properties of actions and improve the model’s representative ability. The model works in a self-augmented fashion, where no external resources are required.
  \item We design the global-local adversarial training scheme to enhance the model’s robustness to irrelevant noises, with respect to video-level prediction and segment-level action-background discrimination. 
  \item We lay out the network architecture to integrate separate modules into a unified framework, which is optimized in an end-to-end fashion. Extensive experiments on challenging untrimmed video datasets show promising results of ActShufNet over the state-of-the-arts.
\end{itemize}

The rest of this article is organized as follows. We review the related work in Section II and introduce the details of the proposed method in Section III. The results of experimental evaluation are reported in Section IV, followed by conclusions in Section V.

\section{Related Work}
\textbf{Action recognition} is aimed to determine the categories of human actions in a trimmed video. Earlier methods extract hand-crafted features, such as Improved Dense Trajectory (iDT)~\cite{idt1,idt2}, consisting of MBH, HOF and HOG features extracted along dense trajectories. Recently, with the development of deep learning, various learning based methods have been proposed, two-stream networks~\cite{two-stream1} learn both spatial and temporal features by operating network on single frame and stacked optical flow field respectively using 2D Convolutional Neural Network. C3D~\cite{c3d} uses 3D convolutional networks to capture both spatial and temporal information directly to learn discriminative features. I3D~\cite{i3d} is exploited to use a 3D version of Inception network~\cite{IoffeS15} under the two-stream architecture. Wang et al.~\cite{tsn} developed a temporal segment network to perform space sparse sampling and fuse the temporal results. There are also having methods using recurrent neural networks to model temporal information, such as LSTM~\cite{lstm}.

\begin{figure*}[htb]
\centering
\includegraphics[width=1\linewidth]{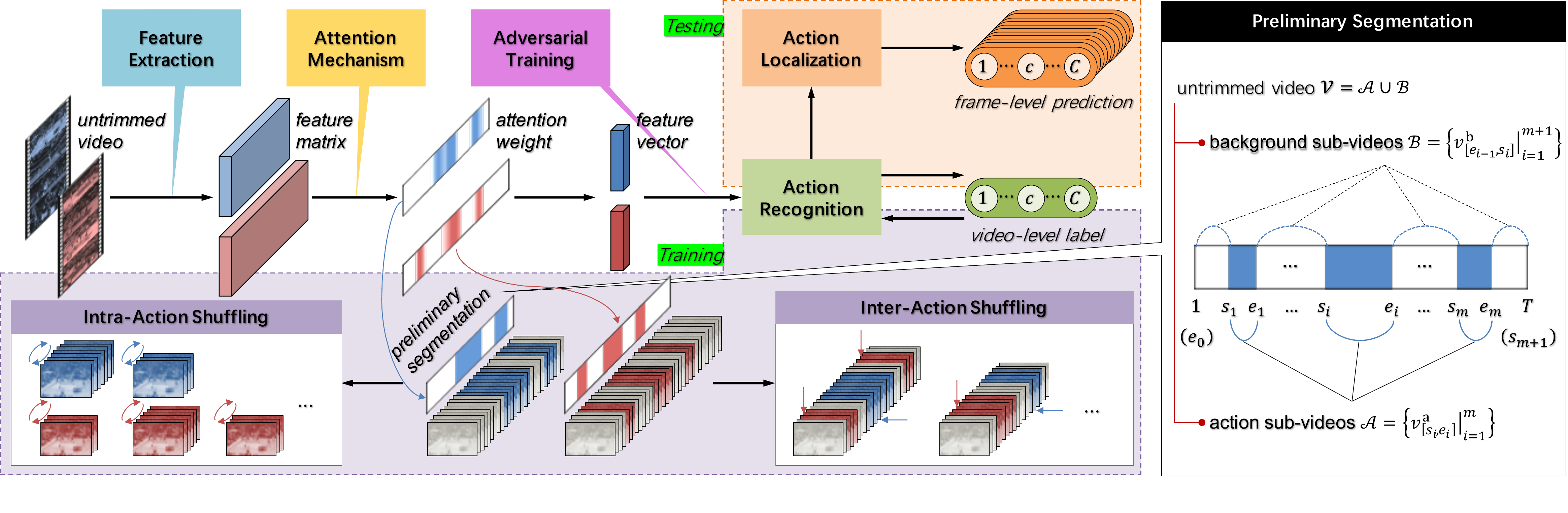}
\caption{Detailed framewrok of ActShufNet. In the training stage, ActShufNet starts from \textit{Feature Extraction}. The extracted features are then fed into \textit{Attention Mechanism} module to obtaiin compact feature representations. Then the action recognition task is optimized with an \textit{Adversarial Training} scheme. In the testing stage, the localization results are obtained.}
\label{fig2}
\end{figure*}

\textbf{Temporal action localization} is aimed to identify the temporal intervals which contain target actions. Previous works mainly focus on designing hand-crafted feature representations to classify the sliding windows~\cite{slidingwindow}. Recently, fully supervised action localization methods leverage the ideology of object detection to obtain improved localization results. SSAD~\cite{ssad} utilizes the 1D temporal convolutional layers to directly detect action instances in untrimmed videos. SSN~\cite{ssn} proposes to utilize a structured temporal pyramid to model the temporal structure of the action instances. S-CNN~\cite{scnn} utilizes multi-stage CNNs to learn hierachical feature representations. BSN~\cite{BSN} utilizes a multi-stage local to global fashion to generate temporal proposals. Besides the fully supervised methods, weakly supervised methods have also been extensively studied, which can be categorized into two classes. Top-down methods (\textit{e.g.}, UntrimmedNets~\cite{untrimmednets}, W-TALC~\cite{w-talc}, 3C-Net~\cite{3c}, CMCS~\cite{CMCS}, BasNet~\cite{BasNet}) learn a video-level classifier and then generate the frame activation score to localize actions. Top-down methods directly learn the temporal attention from videos and optimize the attention with video classification task. TSRNet~\cite{TSRNet} takes advantage of self-attention mechanism and transfer learning and integrates them to obtain precise temporal intervals in untrimmed videos. Autoloc~\cite{autoloc} is proposed to train the boundary predictor with an outerinner-contrastive loss to directly predict the temporal boundary of each action instance. STPN~\cite{STPN} adds a sparse constraint to encourage the action sparsity. BM~\cite{nguyen2019weakly} penalizes the background features and proposes a clustering loss to seperate actions and backgrounds. DGAM~\cite{DGAM} proposes to model the class-agnostic frame-wise probability conditioned on the frame attention using conditional VAE.

\textbf{Self-supervised learning} is one type of techniques that learn representations by solving labourious annotation tasks, where the pseudo supervision signals can be obtained. Self-supervised learning for video analysis aims to learn the motion representations from the unlabeled data by solving the pretext tasks. Wang et al.~\cite{ss1} exploits different self-supervised approaches to learn representations invariant to inter-instance and intra-instance variations among object patches, which are extracted from unlabeled videos using motion cues. In~\cite{ss2}, the chronological order of frames in the video are exploited to learn a robuts temporal features. Similarly, Luo and Wang et al.~\cite{ss,ss3} propose to learn the video representation by predicting motion flows. Inspired by~\cite{ss2}, we also explore the chronological order of action frames to learn precise features. To the best of our knowledge, this is the first attempt that integrate self-supervised learning with weakly supervised action localization.

\textbf{Adversarial learning} has been utilized in computer vision fields to reconstruct target features. In general, it is used to improve the robustness of model. Goodfellow et al. propose an adversarial method namely FGSM~\cite{fgsm} and FGM~\cite{fgm}, which make the direction of pertubation is along the direction of gradient improvement. Madry et al.~\cite{pgd} propose to use projected gradient descent (PGD) to address the internal maximum problem. Since the seminal work by Goodfellow et al.~\cite{gan1} in 2014, a series of GAN family methods have been proposed for a wide variety of problems, which are based on adversarial process corresponding to a minimax two-player game. By means of adversarial learning, we construct the adversary between actions and backgrounds to discriminate them precisely.

\section{Proposed Method}
In this section, we present the framework of intra/inter-action shuffling, i.e., ActShufNet, which is illustrated in Fig.~\ref{fig2}. As a weakly supervised learning model, ActShufNet learns from untrimmed videos and the corresponding video-level labels in the training stage, and predicts frame-level labels of untrimmed videos in the testing stage. For a video $\mathcal{V}=\{f_t|_{t=1}^T\}$ of $T$ frames/snippets, we follow the two-stream standard practice and extract the RGB or optical flow video features $\bm{\mathrm{X}}=[\bm{\mathrm{x}}_t|_{t=1}^T]\in \mathbb{R}^{d\times T}$ with a pre-trained feature extraction model, where $\bm{\mathrm{x}}_t \in \mathbb{R}^d$ is the feature vector of the $t$-th frame/snippet and $d$ is the feature dimension. Without loss of generality, we use frame-wise feature extraction, though the proposed method is also applicable to snippet-wise features. The video-level label is denoted as $\bm{\mathrm{y}}=[y_c|_{c=1}^{C+1}] \in \mathbb{R}^{C+1}$, where $C$ is the number of actions of interest and the $(C+1)$-th class corresponds to background. Given $\bm{\mathrm{X}}$, the model outputs the non-overlapping action instances as $\{(s_i,e_i,\bm{\mathrm{p}}_i)|_{i=1}^m\}$, where $s_i$, $e_i$, and $\bm{\mathrm{p}}_i$ represent the start time, end time, and predictive class label of the $i$-th action instance respectively, and $m$ is the number of identified action instances. 

\subsection{Attention-Based Representation Learning}
Untrimmed videos of variable length $T$ bring about variable-sized feature matrices, which are extremely inconvenient to process. Therefore, we leverage the attention-based mechanism to integrate the frame-level descriptions, and obtain fixed-sized compact representations. 

The attention-based representation learning module aims at deriving attention vector $\bm{\mathrm{\lambda}}=[\lambda_t|_{t=1}^T] \in \mathbb{R}^T$ of an untrimmed video by optimizing the action recognition task. The attention weight $\lambda_t \in [0,1]$ indicates the contribution of the $t$-th frame in identifying an action. Using $\bm{\mathrm{\lambda}}$ to perform attention-weighted temporal average pooling over frames, we arrive at the fixed-sized action representation $\bm{\mathrm{x}}_{[\mathrm{start,end}]}^{\mathrm{a}} \in \mathbb{R}^d$ of video segment $v_{[\mathrm{start,end}]}=\{f_t|_{t=\mathrm{start}}^{\mathrm{end}}\}$ between an arbitrary interval $[\mathrm{start,end}] (1 \leq \mathrm{start} < \mathrm{end} \leq T)$ as follows.

\begin{equation}
\bm{\mathrm{x}}_{[\mathrm{start,end}]}^{\mathrm{a}} = \frac{\sum_{t=\mathrm{start}}^{\mathrm{end}}\lambda_t\bm{\mathrm{x}}_t}{\sum_{t=\mathrm{start}}^{\mathrm{end}}\lambda_t}.
\label{equ1}
\end{equation}
Specifically for the entire video $\mathcal{V}$, the action feature is computed as $\bm{\mathrm{x}}^{\mathrm{a}} = \bm{\mathrm{x}}_{[1,T]}^{\mathrm{a}}$.

Similarly, $(1-\lambda_t)$ can be regarded as the confidence that no actions are taking place in frame $t$ and we can calculate the background representation $\bm{\mathrm{x}}_{[\mathrm{start,end}]}^{\mathrm{b}} \in \mathbb{R}^d$ as follows.
\begin{equation}
\bm{\mathrm{x}}_{[\mathrm{start,end}]}^{\mathrm{b}} = \frac{\sum_{t=1}^{T}(1-\lambda_t)\bm{\mathrm{x}}_t}{\sum_{t=1}^{T}(1-\lambda_t)}.
\end{equation}
The background feature for the entire video $\mathcal{V}$ is $\bm{\mathrm{x}}^{\mathrm{b}}=\bm{\mathrm{x}}_{[1,T]}^{\mathrm{b}}$

For action recognition, the classification loss should encourage the discriminative ability on both action and background.
\begin{equation}
\begin{split}
\mathcal{L}_{\mathrm{CLS}} & = \mathcal{L}_{\mathrm{a}} + \alpha\mathcal{L}_{\mathrm{b}}\\
& = L_{\mathrm{ce}}(p_{\mathrm{cls}}(\bm{\mathrm{x}}^{\mathrm{a}}),\bm{\mathrm{y}}) + \alpha L_{\mathrm{ce}}(p_{\mathrm{cls}}(\bm{\mathrm{x}}^{\mathrm{b}}),\bm{\mathrm{y}}^{\mathrm{b}}).
\end{split}
\label{equ3}
\end{equation}
where $L_{\mathrm{ce}}(\bm{\mathrm{p}},\bm{\mathrm{y}}) = -\bm{\mathrm{y}}^{\top}\mathrm{log}(\bm{\mathrm{p}})$ is the cross-entropy loss, $p_{\mathrm{cls}}(\cdot)$ is the probability output of the action recognition module, and $\bm{\mathrm{y}}^{\mathrm{b}} \in [0,...,0,1]$ is the background label.

For action localization, the temporal class activation maps (TCAM) are utilized to locate the key frames that trigger the video-level label. Given a video and the video-level label $\bm{\mathrm{y}}$, the TCAM is computed as:
\begin{equation}
\hat{\lambda}_t^{\mathrm{a}} = G(\sigma_{\mathrm{s}}) * \frac{\sum_{y_c\neq 0}\mathrm{exp}(\bm{\mathrm{w}}_c\bm{\mathrm{x}}_t)}{\sum_{c=1}^{C+1}\mathrm{exp}(\bm{\mathrm{w}}_c\bm{\mathrm{x}}_t)}.
\end{equation}

\begin{equation}
\hat{\lambda}_t^{\mathrm{b}} = G(\sigma_{\mathrm{s}}) * \frac{\sum_{c=1}^C\mathrm{exp}(\bm{\mathrm{w}}_c\bm{\mathrm{x}}_t)}{\sum_{c=1}^{C+1}\mathrm{exp}(\bm{\mathrm{w}}_c\bm{\mathrm{x}}_t)}.
\end{equation}
where $\bm{\mathrm{w}}_c$ denotes the parameters of the classification module for class $c$. $G(\sigma_{\mathrm{s}})$ is a Gaussian smooth filter with standard deviation $\sigma_{\mathrm{s}}$, and $*$ represents convolution. The TCAM $\hat{\lambda}_t^{\mathrm{a}}$ and $\hat{\lambda}_t^{\mathrm{b}}$ are expected to be consistent with the attention $\hat{\lambda}_t$, and thus used to refine the attention via the self-guided loss. 
\begin{equation}
\mathcal{L}_{\mathrm{guide}} = \frac{1}{T}\sum_{t=1}^{T+1}|\lambda_t-\hat{\lambda}_t^{\mathrm{a}}|+|\lambda_t-\hat{\lambda}_t^{\mathrm{b}}|.
\end{equation}
Based on the attention $\bm{\mathrm{\lambda}}$, we can preliminarily separate action and background segments in an untrimmed video. Note that the separation is coarse-grained compared with the finer-grained action localization. However, the action and background can provide new perspectives for robust representation learning. As illustrated in the right part of Figure~\ref{fig2}, a video $\mathcal{V}$ can be segmented into $m$ action sub-videos $\mathcal{A}=\{v_{[s_i,e_i]}^{\mathrm{a}}|_{i=1}^m\}$ and $(m+1)$ background sub-videos $\mathcal{B}=\{v_{[e_{i-1},s_i]}^{\mathrm{b}}|_{i=1}^{m+1}\}$. Specifically, $e_0 = 1$ and $s_{m+1} = T$ are the first and last frame of the video respectively.

\subsection{Intra-Action Shuffling}
As discussed earlier, the frame order within an action segment is crucial to understand the semantics. In order to improve the representative ability of the attention module, we perform intra-action shuffling and develop a self-supervised order restoring task. To be specific, we sample non-overlapping clips from a preliminarily segmented action sub-video, and shuffle them to a random order. Using the original order as self-supervision, we learn a clip order prediction model based on the clips' attention-weighted features.

Formally, from each action sub-video $v_{[s_i,e_i]}^{\mathrm{a}} \in \mathcal{A}$, we uniformly sample $N$ fixed-sized clips with intervals in between, denoted as $\{v_{[s_{i,k},e_{i,k}]}\lvert_{k=1}^N\} \in v_{[s_i,e_i]}^{\mathrm{a}}$ where $s_i \leq s_{i,k} < e_{i,k} \leq e_i$. In this way, the scattered clips can describe different phases of the action, and meanwhile they share less resemblance with each other. According to Eq.~(\ref{equ1}), the feature vector of each clip is $\bm{\mathrm{x}}_{[s_{i,k},e_{i,k}]}^{\mathrm{a}}$. The sampled $N$ clips are randomly shuffled and organized into a tuple to form the input data, with their original order serving as the target. We formulate order prediction as a classification task, which outputs the probability estimation of the input clip features over different orders. The order prediction module is implemented with the multi-layer perceptron (MLP) structure. The clip features are firstly pairwise concatenated. Each concatenated pair is fed into the ReLU function to obtain a relation vector, i.e., $\bm{\mathrm{r}}_{kj}$, which captures the relation between the two clips. The relation vectors are further concatenated and go through a FC layer with softmax to arrive at the predicted order $\bm{\mathrm{p}}_{\mathrm{ord}}$. The order prediction operations are formulated as follows.
\begin{equation}
\bm{\mathrm{r}}_{kj} = \mathrm{ReLU}(\bm{\mathrm{W}}_1(\bm{\mathrm{x}}_{[s_{i,k},e_{i,k}]}^{\mathrm{a}}\lVert\bm{\mathrm{x}}_{[s_{i,j},e_{i,j}]}^{\mathrm{a}})+\bm{\mathrm{b}}_1).
\end{equation}
\begin{equation}
\bm{\mathrm{p}}_{\mathrm{ord}} = \mathrm{softmax}(\bm{\mathrm{W}}_2(\lVert_{k\prec j}\bm{\mathrm{r}}_{kj})+\bm{\mathrm{b}}_2).
\end{equation}
where $\lVert$ is the concatenation operation of vectors, $k \prec j$ means clip $k$ is in front of clip $j$, and $\bm{\mathrm{W}}_1$, $\bm{\mathrm{b}}_1$, $\bm{\mathrm{W}}_2$, and $\bm{\mathrm{b}}_2$ are parameters of linear transformations.

The order prediction module is optimized with the ordering loss based on cross-entropy function as follows.
\begin{equation}
\mathcal{L}_{\mathrm{intra}} = L_{\mathrm{ce}}(\bm{\mathrm{p}}_{\mathrm{ord}},\bm{\mathrm{y}}^{\mathrm{ord}}).
\end{equation}
where $\bm{\mathrm{y}}^{\mathrm{ord}} \in \mathbb{R}^{N!}$ is the original order. As we can see, the number of all possible orders, i.e., $N!$, overgrows with the increase in the number of clips. For the sake of efficiency, we set $N=5$, which yields $5!=120$ orders.

Intra-action shuffling develops the self-supervised order prediction task, which implicitly benefits representation learning. The ordering loss encourages the attention-based video features to grasp the order-sensitive information within actions, so that the dynamical coherence can be well embedded to enhance the representative ability.

\subsection{Inter-Action Shuffling}
Different from the sensitivity to intra-action frame order, the action is relatively location-insensitive. In other words, re-locating an action as a whole without altering its inner contents will not affect the semantics. Based on the location-insensitive property, we develop the inter-action shuffling strategy to create new training videos. To be specific, we randomly select several action segments from videos of an identical class, and concatenate them into a new video. It is reasonable that the shared class label is still applicable to the new video. Therefore, the extended videos are naturally attached with video-level labels, and can be safely used as additional training data.

Let $\mathcal{T} = \{(\mathcal{V}^{(l)},\bm{\mathrm{y}}^{(l)})|_{l=1}^L\}$ denotes the training dataset comprised of $L$ videos with the corresponding video-level labels. Given a predicted action instance $v_{[s_i,e_i]}^{a(l)} \in \mathcal{A}^{(l)} \subset \mathcal{V}^{(l)}$ using attention-based segmentation, we inflate its boundary slightly to obtain the outer-boundary action sub-video $v_{[s_i-\Delta,e_i+\Delta]}^{(l)}$, where $\Delta$ is the inflation interval. From the videos of a specific action class $\tilde{\bm{\mathrm{y}}}$, we randomly select the outer-boundary action sub-videos, and generate a new video $\tilde{\mathcal{V}}=\{v_{[s_i-\Delta,e_i+\Delta]}^{(l)}|l \in [1,L],\bm{\mathrm{y}}^{(l)}=\tilde{\bm{\mathrm{y}}}, i \in [1,m^{(l)}]\}$ with video-level label $\tilde{\bm{\mathrm{y}}}$, where $m^{(l)}$ is the number of predicted action instances in $\mathcal{V}^{(l)}$. Note that, for the generated video, there is no need for feature extraction from scratch. The features $\tilde{\bm{\mathrm{X}}}$ of generated video $\tilde{\bm{\mathrm{V}}}$ can be conveniently obtained by simply concatenating the corresponding frame feature vectors, based on which the attention-based representations are also learned. In this way, an additional training dataset $\tilde{\mathcal{T}} = \{(\tilde{\mathcal{V}}^{(l)},\tilde{\bm{\mathrm{y}}}^{(l)})|_{l=1}^{\tilde{L}}\}$ is created, where $\tilde{L}$ is the number of generated videos.

To correctly identify action and background in the generated video, the inter-action shuffling module is optimized with the classification loss similar to Eq.~(\ref{equ3}) as follows.
\begin{equation}
\mathcal{L}_{\mathrm{inter}} = L_{\mathrm{ce}}(p_{\mathrm{cls}}(\tilde{\bm{\mathrm{x}}}^{\mathrm{a}}),\tilde{\bm{\mathrm{y}}})+\alpha L_{\mathrm{ce}}(p_{\mathrm{cls}}(\tilde{\bm{\mathrm{x}}}^{\mathrm{b}}),\tilde{\bm{\mathrm{y}}}^{\mathrm{b}}).
\label{equ10}
\end{equation}

\begin{table}[tbp]
\small
\centering
\begin{tabular}{l|l|c}
\hline
Datasets&Method&Accuracy\\
\hline
\multirow{5}*{THUMOS14}
&UNets (Wang et al., 2017)~\cite{untrimmednets}& 82.2 \\
&W-TALC (Paul et al., 2018)~\cite{w-talc}&85.6 \\
&TSRNet (Zhang et al., 2019)~\cite{TSRNet}& 87.1 \\
&PreTrimNet (Zhang et al., 2019)~\cite{pretrimnet}& 89.2\\
&ActShufNet& \textbf{92.8} \\
\hline \hline
\multirow{2}*{ActivityNet1.2}
&W-TALC (Paul et al., 2018)~\cite{w-talc}& 93.2 \\
&ActShufNet& \textbf{93.8} \\
\hline \hline
\multirow{3}*{ActivityNet1.3}
&TSRNet (Zhang et al., 2019)~\cite{TSRNet}& 91.2 \\
&PreTrimNet (Zhang et al., 2019)~\cite{pretrimnet}& 93.3\\
&ActShufNet& \textbf{93.4} \\
\hline
\end{tabular}
\caption{Comparison of action recognition results on THUMOS14 and ActivityNet (1.2 and 1.3).}
\label{table:table1}
\end{table}

Inter-action shuffling benefits the representation learning in a weakly supervised learning fashion. On one hand, the training dataset can be effectively expanded by generating additional video-level labeled videos, and meanwhile the problem of data imbalance can be naturally solved by adaptive generation according to data distribution. On the other hand, inter-action shuffling introduces more variety in each generated video, and provides a more challenging auxiliary training dataset to shape a more robust attention model.

\subsection{Global-Local Adversarial Training}
In each untrimmed video, action and background segments bear resemblance to some extent in visual and motional aspects. Therefore, action and background can be easily confused with each other, which jeopardizes the localization performance. Recent studies indicate that adversarial training is effective to enhance the model’s tolerance to irrelevant noises, by adding small perturbations to the inputs. To further improve the discriminative ability of the classification module, we develop the global-local adversarial training scheme to achieve robust learning performance. 

To be specific, global adversarial training focuses on the robustness of video-level prediction, which is formulated as follows by allowing perturbations to the video-level action and background feature vectors in the classification loss in Eq.~(\ref{equ3}) with video-level labels.
\begin{equation}
\begin{split}
\mathcal{L}_{\mathrm{global}} &= \max_{\delta^{\mathrm{a}}}L_{\mathrm{ce}}(p_{\mathrm{cls}}(\bm{\mathrm{x}}^{\mathrm{a}}+\bm{\mathrm{\delta}}^{\mathrm{a}}),\bm{\mathrm{y}})\\
&+ \alpha\max_{\delta^{\mathrm{b}}}L_{\mathrm{ce}}(p_{\mathrm{cls}}(\bm{\mathrm{x}}^{\mathrm{b}}+\bm{\mathrm{\delta}}^{\mathrm{b}}),\bm{\mathrm{y}}^{\mathrm{b}}).
\end{split}
\label{equ11}
\end{equation}
where $\delta^{\mathrm{a}}$ and $\delta^{\mathrm{b}}$ are perturbations imposed on $\bm{\mathrm{x}}^{\mathrm{a}}$ and $\bm{\mathrm{x}}^{\mathrm{b}}$, respectively. $\max_{\bm{\mathrm{\delta}}}L_{\mathrm{ce}}(\cdot)$ finds the perturbation that maximizes the loss function and is most likely to fool the classifier. Global adversarial training encourages consistence between video-level prediction and supervision under perturbation, and makes the model less affected by irrelevant contents. 

Different from the global scheme, local adversarial training focuses on segments instead of the entire video. It aims to optimize the action-background separation by maximizing the distinction between adjacent action and background. Formally, for action segment $v_{[s_i,e_i]}^{\mathrm{a}}\in \mathcal{V}$, the adjacent background segments are $v_{[e_{i-1},s_i]}^{\mathrm{b}}$ and $v_{[e_i,s_{i+1}]}^{\mathrm{b}}$. Local adversarial training encourages different predictions of each adjacent action-background pair, which is formulated as follows.
\begin{equation}
\begin{split}
&\mathcal{L}_{\mathrm{local}} = \sum_{i=1}^m \max_{\bm{\mathrm{\delta}}^{\mathrm{a}}} \\
&-L_{\mathrm{ce}}(p_{\mathrm{cls}}(\bm{\mathrm{x}}_{[e_{i-1},s_i]}^{\mathrm{a}}+\bm{\mathrm{\delta}}_{[e_{i-1},s_i]}^{\mathrm{b}}),p_{\mathrm{cls}}(\bm{\mathrm{x}}_{[s_i,e_i]}^{\mathrm{a}}+\bm{\mathrm{\delta}}_{[s_i,e_i]}^{\mathrm{a}}) \\
&-L_{\mathrm{ce}}(p_{\mathrm{cls}}(\bm{\mathrm{x}}_{[s_i,e_i]}^{\mathrm{a}}+\bm{\mathrm{\delta}}_{[s_i,e_i]}^{\mathrm{a}}),p_{\mathrm{cls}}(\bm{\mathrm{x}}_{[e_i,s_{i+1}]}^{\mathrm{a}}+\bm{\mathrm{\delta}}_{[e_i,s_{i+1}]}^{\mathrm{b}}).
\end{split}
\label{equ12}
\end{equation}
Note that, for all the segments, we use the action representation in Eq.~(\ref{equ1}) for unified comparison. 

Directly calculating $\bm{\mathrm{\delta}}$ that maximizes the loss function is infeasible. In this paper, we follow the fast gradient sign method (FGSM) to obtain the perturbation as follows.
\begin{equation}
\bm{\mathrm{\delta}} = \epsilon \mathrm{sign}(\triangledown_{\bm{\mathrm{x}}}f(\bm{\mathrm{x}})).
\label{equ13}
\end{equation}
where $\epsilon$ is a pre-defined hyper-parameter, and $f(\bm{\mathrm{x}})$ is the loss function w.r.t. $\bm{\mathrm{x}}$, i.e., $f(\cdot)=L_{\mathrm{ce}}(\cdot)$ for Eq.~(\ref{equ11}) and $f(\cdot)=-L_{\mathrm{ce}}(\cdot)$ for Eq.~(\ref{equ12}).

The global-local adversarial training loss can be defined as:
\begin{equation}
\mathcal{L}_{\mathrm{adv}} = \mathcal{L}_{\mathrm{global}} + \beta\mathcal{L}_{\mathrm{local}}.
\label{equ14}
\end{equation}

Finally, we arrive at the overall loss of ActShufNet.
\begin{equation}
\mathcal{L} = \mathcal{L}_{\mathrm{adv}} + \eta\mathcal{L}_{\mathrm{intra}} + \theta\mathcal{L}_{\mathrm{inter}} + \gamma\mathcal{L}_{\mathrm{guide}}.
\label{equ15}
\end{equation}

In the training stage, ActShufNet is optimized by minimizing the overall loss in Eq. (\ref{equ15}). In the testing stage, the well-trained ActShufNet makes frame-level class activation prediction to achieve temporal localization.

\begin{table*}[htb]
\small
\centering
\begin{tabular}{l|rrrrrrrrr}
\hline
\multirow{2}*{Method}&\multicolumn{9}{c}{mAP@IoU (\%)}\\
\cline{2-10}
 &0.1&0.2&0.3&0.4&0.5&0.6&0.7&0.8&0.9\\
\hline
\multirow{4}*{}SCNN (Shou et al., 2016)~\cite{scnn} &47.7&43.5&36.3&28.7&19.0&10.3&5.3&-&-\\
SSN (Zhao et al., 2017)~\cite{ssn}&66.0&59.4&51.9&41.0&29.8&-&-&-&-\\
TAL-Net (chao et al., 2018)~\cite{chao}&59.8&57.1&53.2&48.5&42.8&\bf{33.8}&\bf{20.8}&-&-\\
BSN (Lin et al., 2018)~\cite{BSN}&-&-&53.5&45.0&36.9&28.4&20.0&-&-\\
P-GCN (Zeng et al., 2019)~\cite{zeng2019graph}&\bf{69.5}&\bf{67.8}&\bf{63.6}&\bf{57.8}&\bf{49.1}&-&-&-&-\\
SF-Net (Ma et al., 2020)~\cite{SFNet}&68.7&-&54.5&-&34.4&-&16.7&-&-\\
\hline \hline
\multirow{16}*{}UNets (Wang et al., 2017)~\cite{untrimmednets}&44.4&37.7&28.2&21.1&13.7&-&-&-&-\\
STPN (Nguyen et al., 2018)~\cite{STPN}&45.3&38.8&31.1&23.5&16.2&9.8&5.1&2.0&0.3\\
AutoLoc (Shou et al., 2018)~\cite{autoloc}&-&-&35.8&29.0&21.2&13.4&5.8&-&-\\
W-TALC (Paul et al., 2018)~\cite{w-talc}&49.0&42.8&32.0&26.0&18.8&-&6.2&-&-\\
CMCS (Liu et al., 2019)~\cite{CMCS}&53.5&46.8&37.5&29.1&19.9&12.3&6.0&-&-\\
BaS-Net (Lee et al., 2020)~\cite{BasNet}&56.2&50.3&42.8&34.7&25.1&17.1&9.3&3.7&0.5\\
ActShufNet (UNT)&\bf{59.12}&\bf{53.68}&\bf{45.25}&\bf{36.90}&\bf{27.24}&\bf{19.47}&\bf{10.07}&\bf{4.01}&0.39\\
\hline \hline
STPN (Nguyen et al., 2018)~\cite{STPN}&52.0&44.7&35.5&25.8&16.9&9.9&4.3&1.2&0.1\\
W-TALC (Paul et al., 2018)~\cite{w-talc}&55.2&49.6&40.1&31.1&22.8&-&7.6&-&-\\
TSRNet (Zhang et al., 2019)~\cite{TSRNet}&55.9&46.9&38.3&28.1&18.6&11.0&5.59&2.19&0.29\\
TSM (Yu et al., 2019)~\cite{TSM}&-&-&39.5&31.9&24.5&13.8&7.1&-&-\\
CMCS (Liu et al., 2019)~\cite{CMCS}&57.4&50.8&41.2&32.1&23.1&15.0&7.0&-&-\\
BM (Nguyen et al., 2019)~\cite{nguyen2019weakly}&60.4&56.0&46.6&37.5&26.8&17.6&9.0&3.3&0.4\\
3C-Net (Narayan et al., 2019)~\cite{3c}&59.1&53.5&44.2&34.1&26.6&-&8.1&-&-\\
MAAN (Yuan et al., 2019)~\cite{maan}&59.8&50.8&41.1&30.6&20.3&12.0&6.9&2.6&0.2\\
PreTrimNet (Zhang et al., 2019)~\cite{pretrimnet}&57.49&50.73&41.40&32.05&23.09&14.16&7.69&2.33&0.39\\
DGAM (Shi et al., 2020)~\cite{DGAM}&60.0&54.2&46.8&38.2&28.8&19.8&\bf{11.4}&3.6&0.4\\
ActionBytes (Jain et al., 2020)~\cite{actionbytes}&-&-&43.0&35.8&29.0&-&9.5&-&-\\
Deep Metric Learning (Islam et al., 2020)~\cite{DeepMetric}&62.3&-&46.8&-&29.6&-&9.7&-&-\\
BaS-Net (Lee et al., 2020)~\cite{BasNet}&58.2&52.3&44.6&36.0&27.0&18.6&10.4&3.9&0.5\\
TSCN (Zhai et al., 2020)~\cite{tscn}&63.4&57.6&47.8&37.7&28.7&19.4&10.2&3.9&0.7\\
A2CL-PT (Min et al., 2020)~\cite{a2cl}&61.2&56.1&48.1&39.0&30.1&19.2&10.6&\bf{4.8}&\bf{1.0}\\
ActShufNet (I3D)&\bf{63.44}&\bf{57.92}&\bf{48.46}&\bf{40.01}&\bf{31.12}&\bf{22.01}&11.26&4.46&0.50\\
\hline
\end{tabular}
\caption{Comparison of action localization results on THUMOS14. Entries are separated regarding the level of supervision. The partition upon the first double horizontal line represents the fully supervised methods. For weakly-supervised setting, we compare both UntrimmedNet (UNT) features and I3D features, as decipted in the lower two partitions.}
\label{table:table2}
\end{table*}

\begin{table}[htb]
\centering
\begin{tabular}{lcccc}
\hline
\multirow{2}*{Method}&\multicolumn{4}{c}{mAP@IoU (\%)}\\
\cline{2-5}
&0.5&0.75&0.95&Average\\
\hline 
\multirow{5}*{}SSN (Zhao et al., 2017)~\cite{ssn}&41.3&27.0&6.1&26.6\\
\hline \hline
\multirow{7}*{}W-TALC (Paul et al., 2018)~\cite{w-talc}&37.0&12.7&1.5&18.0\\
TSM (Yu et al., 2019)~\cite{TSM}&30.3&19.0&4.5&-\\
CMCS (Liu et al., 2019)~\cite{CMCS}&36.8&22.0&5.6&22.4\\
3C-Net (Narayan et al., 2019)~\cite{3c}&35.4&-&-&21.1\\
CleanNet (Liu et al., 2019)~\cite{Liu_2019_ICCV}&34.5&22.5&4.9&22.2\\
DGAM (Shi et al., 2020)~\cite{DGAM}&41.0&23.5&5.3&24.4\\
TSCN (Zhai et al., 2020)~\cite{tscn}&37.6&23.7&5.7&23.6\\
ActShufNet&\bf{41.2}&\bf{24.9}&\bf{5.9}&\bf{25.0}\\
\hline
\end{tabular}
\caption{Comparison of action localization results on ActivityNet1.2. The partition upon the double horizontal line represents fully supervised methods, belows are weakly supervised methods. The column Average means the average mAP at IoU thresholds 0.5:0.05:0.95.}
\label{table:table3}
\end{table}

\begin{table}[t]
\centering
\begin{tabular}{lcccc}
\hline
\multirow{2}*{Method}&\multicolumn{4}{c}{mAP@IoU (\%)}\\
\cline{2-5}
&0.5&0.75&0.95&Average\\
\hline
\multirow{5}*{}SSN (Zhao et al., 2017)~\cite{ssn}&39.12&23.48&5.49&23.98\\
BSN (Lin et al., 2018)~\cite{BSN}&\textbf{52.50}&\textbf{33.53}&\textbf{8.85}&\textbf{33.72}\\
P-GCN (Zeng et al., 2019)~\cite{zeng2019graph}&42.90&28.14&2.47&31.11\\
\hline \hline
\multirow{7}*{}STPN (Nguyen et al., 2018)~\cite{STPN}&29.3&16.9&2.6&-\\
TSRNet (Zhang et al., 2019)~\cite{TSRNet}&33.1&18.7&3.32&21.78\\
TSM (Yu et al., 2019)~\cite{TSM}&30.3&19.0&4.5&-\\
CMCS (Liu et al., 2019)~\cite{CMCS}&34.0&20.9&5.7&21.2\\
BM (Nguyen et al., 2019)~\cite{nguyen2019weakly}&36.4&19.2&2.9&-\\
MAAN (Yuan et al., 2019)~\cite{maan}&33.7&21.9&5.5&-\\
PreTrimNet (Zhang et al., 2019)~\cite{pretrimnet}&34.8&20.9&5.3&22.5\\
BaS-Net (Lee et al., 2020)~\cite{BasNet}&34.5&22.5&4.9&22.2\\
A2CL-PT (Min et al., 2020)~\cite{a2cl}&\bf{36.8}&22.0&5.2&22.5\\
TSCN (Zhai et al., 2020)~\cite{tscn}&35.3&21.4&5.3&21.7\\
ActShufNet&36.3&\bf{23.5}&\bf{5.8}&\bf{23.6}\\
\hline
\end{tabular}
\caption{Comparison of action localization results on ActivityNet1.3. The partition upon the double horizontal line represents fully supervised methods, belows are weakly supervised methods. The column Average means the average mAP at IoU thresholds 0.5:0.05:0.95.}
\label{table:table4}
\end{table}

\begin{figure*}[htb]
\centering
\subfigure[Video of $\textit{CricketBowling}$ (top) and $\textit{CricketShot}$ (down) actions.]{\includegraphics[width=1\linewidth]{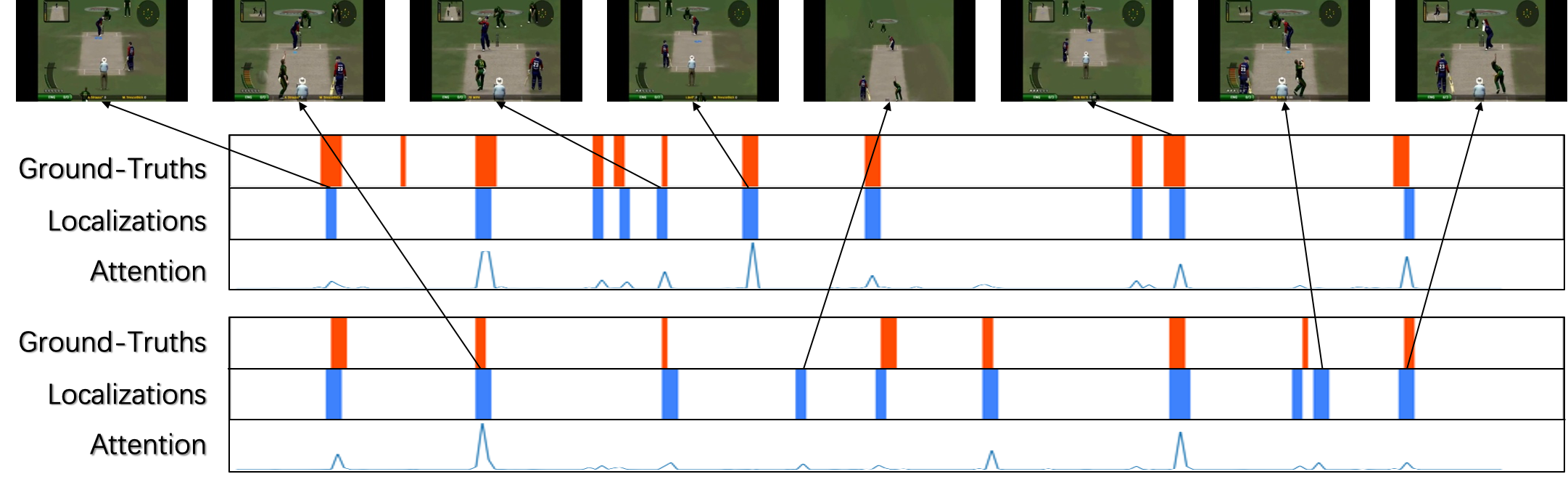}}
\subfigure[Video of $\textit{JavelinThrow}$ action.]{\includegraphics[width=1\linewidth]{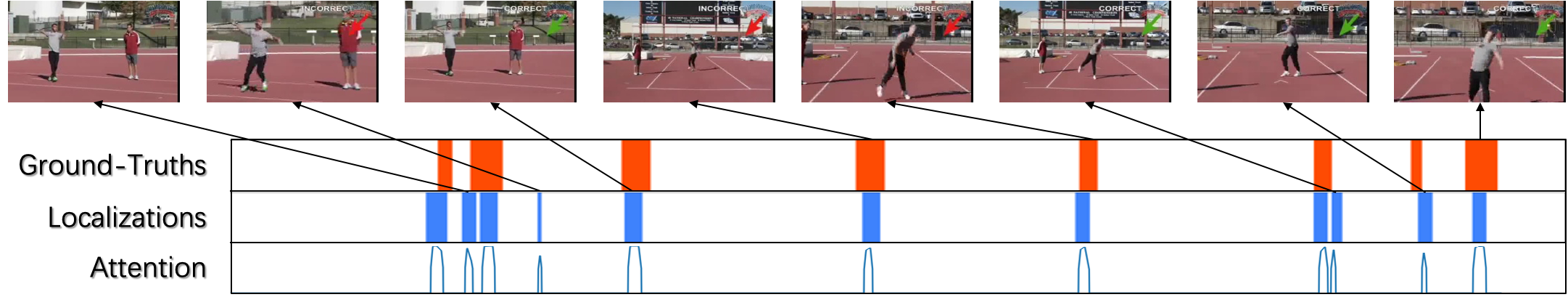}}
\subfigure[Video of $\textit{High\ jump}$ action.]{\includegraphics[width=1\linewidth]{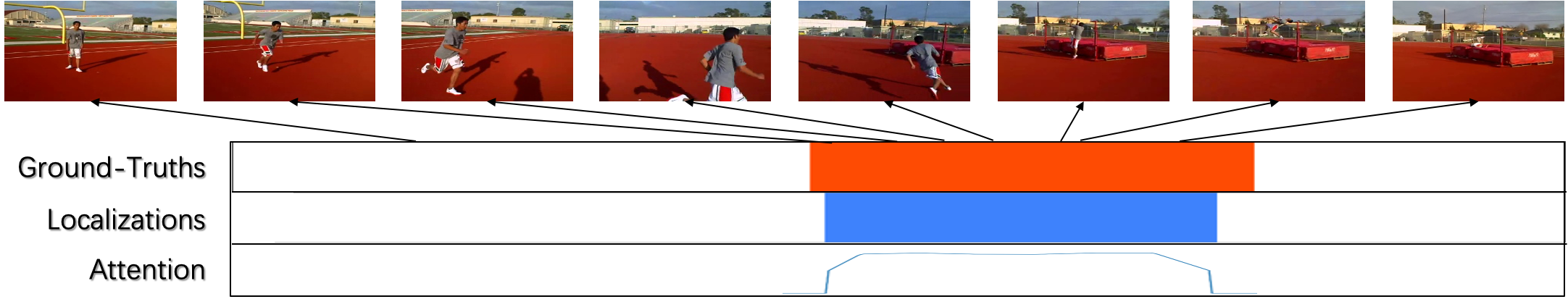}}
\caption{Qualitative results on THUMOS14 (a and b), and ActivityNet (c). The red bars denote the ground-truth. The blue bars denote localization results.}
\label{vis}
\end{figure*}

\section{Experiments}
\subsection{Datasets and Evaluation Metrics}
We evaluate the proposed ActShufNet on three benchmarks, i.e. THUMOS14~\cite{THUMOS14} and two released versions of ActivityNet~\cite{activitynet}. As a weakly supervised method, ActShufNet only has access to the video-level annotations during training.

\textbf{THUMOS14} contains a validation set and a testing set of 1,010 and 1,574 videos, respectively. There are 101 action classes, among which 20 classes are temporally annotated. We focus on the 20-class subset, using the validation set of 200 videos for training and the testing set of 213 videos for evaluation. THUMOS14 is challenging in that it contains videos with multiple actions.

\textbf{ActivityNet} has two released versions, i.e., ActivityNet1.2 and ActivityNet1.3. ActivityNet1.2 contains 100 classes of videos, with 4,819 videos for training, 2,383 for validation, and 2,480 for testing. ActivityNet1.3 is an extension of ActivityNet1.2, which is comprised of 200 activity classes, with 10,024 videos for training, 4,926 for validation, and 5,044 for testing. Since the ground-truth labels for the original testing set are withheld, we adopt the training set for model training and the validation set for testing.

\textbf{Evaluation Metrics.} We follow the standard evaluation protocol and report mean Average Precision (mAP) at different intersection over union (IoU) thresholds. The results are calculated using the benchmark code provided by ActivityNet official codebase\footnote{https://github.com/activitynet/ActivityNet/tree/master/Evaluation}.

\subsection{Implementation Details}
We utilize the two-stream I3D networks pre-trained on Kinetics dataset to extract the traditional two-stream features. For the RGB stream, we perform the center crop of size 224 $\times$ 224. For the optical flow stream, we apply the TV-$L$1 optical flow algorithm. The input to the I3D models are stacks of 16 (RGB or flow) frames sampled at 16 frames per second to obtain two 1024-dimension video features. The model parameters are optimized using the mini-batch stochastic gradient descent with Adam optimizer. The learning rate is set to 1$e$-4 for both RGB and optical flow streams. We also utilize the dropout operations with ratios 0.5 and common augmentation techniques including horizontal flipping, cropping augmentation, et al. We set the parameters $\alpha$, $\beta$, $\epsilon$, $\eta$, $\theta$ and $\gamma$ to 1, 0.01, 0.001, 1, 0.1 and 0.1, respectively. Our algorithm is implemented in PyTorch.

\subsection{Results}

\begin{figure}[tb]
\centering
{\includegraphics[width=0.453\linewidth]{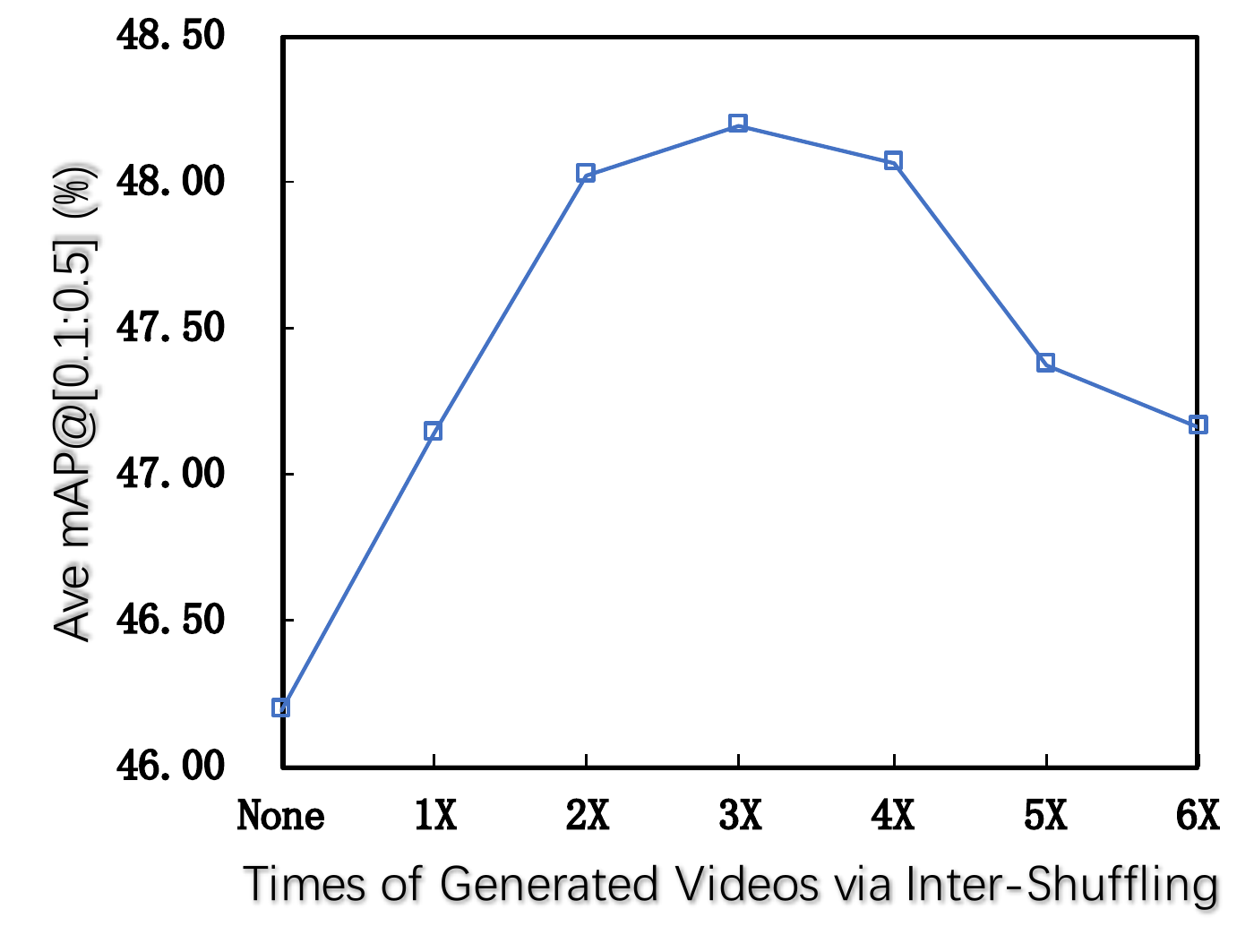}
}
\quad
{\includegraphics[width=0.453\linewidth]{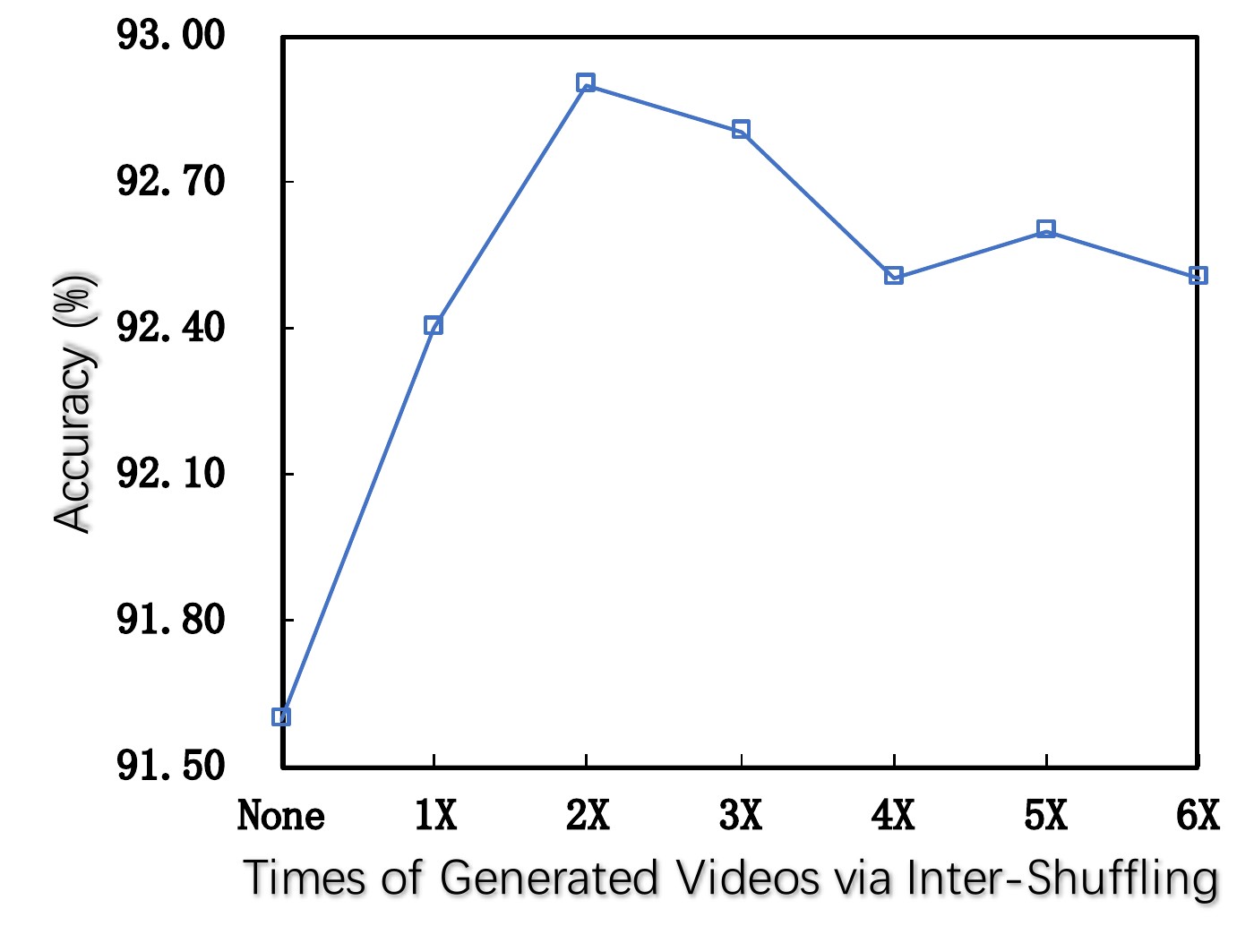}
}
\caption{Action recognition (left) and localization (right) results on THUMOS14 w.r.t. the number of generated videos via inter-shuffling. ``*X" means * times of original number of training videos.}
\label{fig3}
\end{figure}

\textbf{Hyperparameter study.} To investigate the effect of training set augmentation via inter-action shuffling, we study the variation of action recognition and localization results on THUMOS14 w.r.t. the number of generated videos. As illustrated in Fig.~\ref{fig3}, the performance improves continuously until 2 to 3 times of auxiliary training videos are leveraged, and degrades afterward. Therefore, we expand the training set by 3 times throughout the following experiments.

\textbf{Action recognition.} We compare the action recognition performance of ActShufNet with the state-of-the-art methods. As shown in TABLE~\ref{table:table1}, ActShufNet remarkably outperforms most of its competitors on all three benchmarks. One exception is that ActShufNet only slightly surpasses PreTrimNet~\cite{pretrimnet} on ActivityNet1.3. However, it is worth noticing that PreTrimNet is based on fine-grained spatio-temporal segmentation with three-stream features, thus is far more complex than ActShufNet.

\begin{table*}[htb]
\small
\centering
\begin{tabular}{ccl|ccccccccc}
\hline
\multirow{2}*{$\mathcal{L}_{\mathrm{adv}}$}&\multirow{2}*{$\mathcal{L}_{\mathrm{intra}}$}&\multirow{2}*{$\mathcal{L}_{\mathrm{inter}}$}&\multicolumn{9}{c}{mAP@IoU (\%)}\\
\cline{4-12}
& & & 0.1&0.2&0.3&0.4&0.5&0.6&0.7&0.8&0.9\\
\hline
-&-&-&56.22 & 50.57 & 40.46 & 32.05 & 21.89 & 12.45 & 6.81 & 2.06 & 0.26\\
\checkmark&-&-&58.91 & 53.14 & 45.98 & 37.33 & 27.92 & 18.99 & 9.98 & 3.17 & 0.35\\
\checkmark&-&\checkmark&60.42&54.30&46.69&37.93&28.20&19.35&10.80&3.47&0.40\\
\checkmark&\checkmark&-&60.66 & 54.63 & 46.70 & 38.71 & 28.91 & 21.02 & 11.14 & 4.26 & 0.47\\
\checkmark&\checkmark&\checkmark&\bf{63.44}&\bf{57.92}&\bf{48.46}&\bf{40.01}&\bf{31.12}&\bf{22.01}&\bf{11.26}&\bf{4.46}&\bf{0.50}\\
\hline
\end{tabular}
\caption{Comparison of action localization results of ActShufNet with different implementations on THUMOS14.}
\label{table:table5}
\end{table*}

\textbf{Action localization.} We evaluate ActShufNet on temporal action localization task, in comparison with both fully and weakly supervised methods. Localization results on THUMOS14 are listed in TABLE~\ref{table:table2}. The compared methods are compared in chronological order. The lower two partitions are grouped by choice of the feature extractor: UntrimmedNet (UNT) and I3D. It is observed that ActShufNet significantly surpasses its weakly supervised counterparts and achieves the highest average mAP at the same level of supervision, regardless of the feature extractor network. It is especially encouraging to see that ActShufNet even achieves comparable results with some fully supervised methods (in upper parts of the tables).

We also evaluate our ActShufNet on ActivityNet1.2 in TABLE~\ref{table:table3}. We see that our method outperforms all other weakly-supervised approaches. Moreover, despite using generative models (i.e., cVAE), our algorithm outperforms DGAM at all IoU thresholds. Experimental results on ActivityNet1.3 are shown in TABLE~\ref{table:table4} to compare our method with more methods. Our model outperforms all weakly-supervised methods with the average mAP, following the fully-supervised method with a small gap.

\textbf{Ablation study.} To validate the effectiveness of key components, we compare the full implementation of ActShufNet with its abridged versions without some of the losses in Eq.~(\ref{equ15}). TABLE~\ref{table:table5} summarizes the action localization results on THUMOS14. We observe that each component is indispensable to achieve accurate results, and the absence of any component will lead to notable performance deterioration. We further illustrate examples of temporal action localization on THUMOS14 ((a) and (b)) and ActivityNet1.3 ((c)), as shown in Fig.~\ref{vis}. As ActivityNet 1.3 is an extension vision of 1.2, we only visualize the results of 1.3 version. The visualization results include (a) a video containing two action classes, (b) short lasting action, and (c) long lasting action. Generally, the multiple actions in a video seems to be similar, which is vulnerable to the boundary noise. In the case (a), the two actions $\textit{CricketBowling}$ and $\textit{CricketShot}$ have overlapping parts along the time axis, our method can also discriminate the actions. In case (b), there are short lasting actions, where actions happen quickly, and our method can catch the key action frames. In case (c), the video contains a complete $\textit{High\ jump}$ action, which last long time in ground-truths, our method can also track and localize the actions. As we can see, the proposed method is an effective indicator that is capable of locating actions of interest in untrimmed videos under different circumstances.

\section{Conclusion}
In this paper, we have proposed a novel self-augmented framework, namely ActShufNet, for action localization in untrimmed videos with video-level weak supervision. Instead of taking untrimmed video as a whole, we focus on sub-videos derived from preliminary segmentation. Based on analysis of the order-sensitive and location-insensitive properties of actions, we design a two-branch network architecture with intra/inter-action shuffling. The former aims to augment the model’s representative ability via inner-video order shuffling, whereas the latter generates new videos to augment the training set by reorganizing the existing action sub-videos. The global-local adversarial training scheme is further presented to ensure model robustness to irrelevant noises. As demonstrated on three challenging untrimmed video datasets, ActShufNet achieves superior performance over the state-of-the-art weakly supervised methods, and is even comparable to some fully supervised counterparts.


%





\ifCLASSOPTIONcaptionsoff
  \newpage
\fi



%


\bibliographystyle{IEEEtran}
\bibliography{ref}
%
\vspace{30 mm}
\begin{IEEEbiography}[{\includegraphics[width=1in,height=1.25in,clip,keepaspectratio]{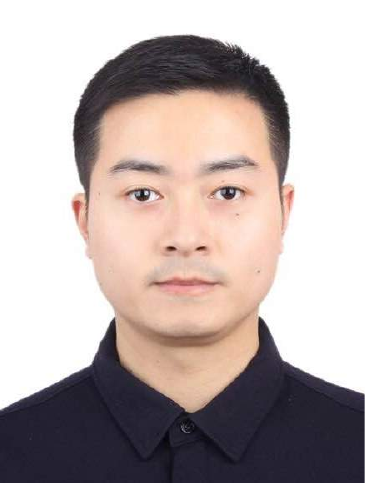}}]{Xiao-Yu Zhang} 
(Senior Member, IEEE) received the B.S. degree in computer science from Nanjing University of Science and Technology, Nanjing, China, in 2005, and the Ph.D. degree in pattern recognition and intelligent systems from the Institute of Automation, Chinese Academy of Sciences, Beijing, China, in 2010. 

He is currently an Associate Professor with the Institute of Information Engineering, Chinese Academy of Sciences, Beijing, China. He has authored or coauthored more than 60 refereed publications in international journals and conferences. His research interests include artificial intelligence, data mining, computer vision, etc.

Dr. Zhang is a Senior Member of the ACM, CCF, and CSIG. His awards and honors include the Silver Prize of Microsoft Cup of the IEEE China Student Paper Contest in 2009, the Second Prize of Wu Wen-Jun AI Science \& Technology Innovation Award in 2016, the CCCV Best Paper Nominate Award in 2017, the Third Prize of BAST Beijing Excellent S\&T Paper Award in 2018, and the Second Prize of CSIG Science \& Technology Award in 2019.
\end{IEEEbiography}

\vspace{-40 mm}
\begin{IEEEbiography}[{\includegraphics[width=1in,height=1.25in,clip,keepaspectratio]{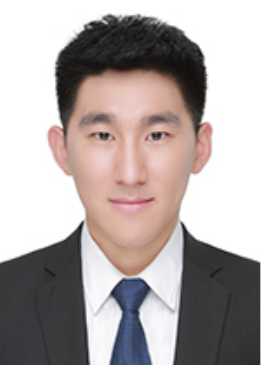}}]{Haichao Shi}
received the B.S. degree in software engineering from Beijing Technology and Business University, Beijing, China, in 2017. He is currently pursuing the Ph.D. degree in cyberspace security with the State Key Laboratory of Information Security, Institute of Information Engineering, Chinese Academy of Sciences, Beijing.

His research interests include pattern recognition, image processing and video content analysis.
\end{IEEEbiography}

\vspace{-40 mm}
\begin{IEEEbiography}[{\includegraphics[width=1in,height=1.25in,clip,keepaspectratio]{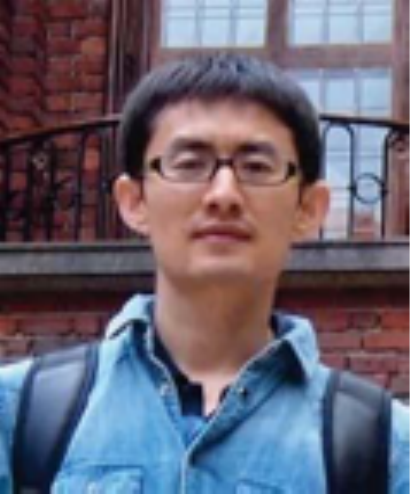}}]{Changsheng Li}
(Member, IEEE) received the B.E. degree from the University of Electronic Science and Technology of China (UESTC), Chengdu, China, in 2008, and the Ph.D. degree in pattern recognition and intelligent system from the Institute of Automation, Chinese Academy of Sciences, Beijing, China, in 2013. 

He was a Research Assistant with The Hong Kong Polytechnic University, Hong Kong, from 2009 to 2010. He worked with IBM Research-China, Beijing, Alibaba Group, Beijing, and UESTC, respectively. He is currently a Professor with the Beijing Institute of Technology, Beijing. He has authored or coauthored more than 40 refereed publications in international journals and conferences, including the IEEE \textsc{Transactions on} \textsc{Pattern} \textsc{Analysis and} \textsc{Machine} \textsc{Intelligence}, the IEEE \textsc{Transcations on} \textsc{Image} \textsc{Processing}, the IEEE \textsc{Transcations on} \textsc{Neural} \textsc{Networks and} \textsc{Learning} \textsc{System}, the IEEE \textsc{Transcations on} \textsc{Computers}, PR, CVPR, AAAI, IJCAI, CIKM, MM, ICMR, etc. His research interests include machine learning, data mining, and computer vision.
\end{IEEEbiography}

\vspace{-40 mm}
\begin{IEEEbiography}[{\includegraphics[width=1in,height=1.25in,clip,keepaspectratio]{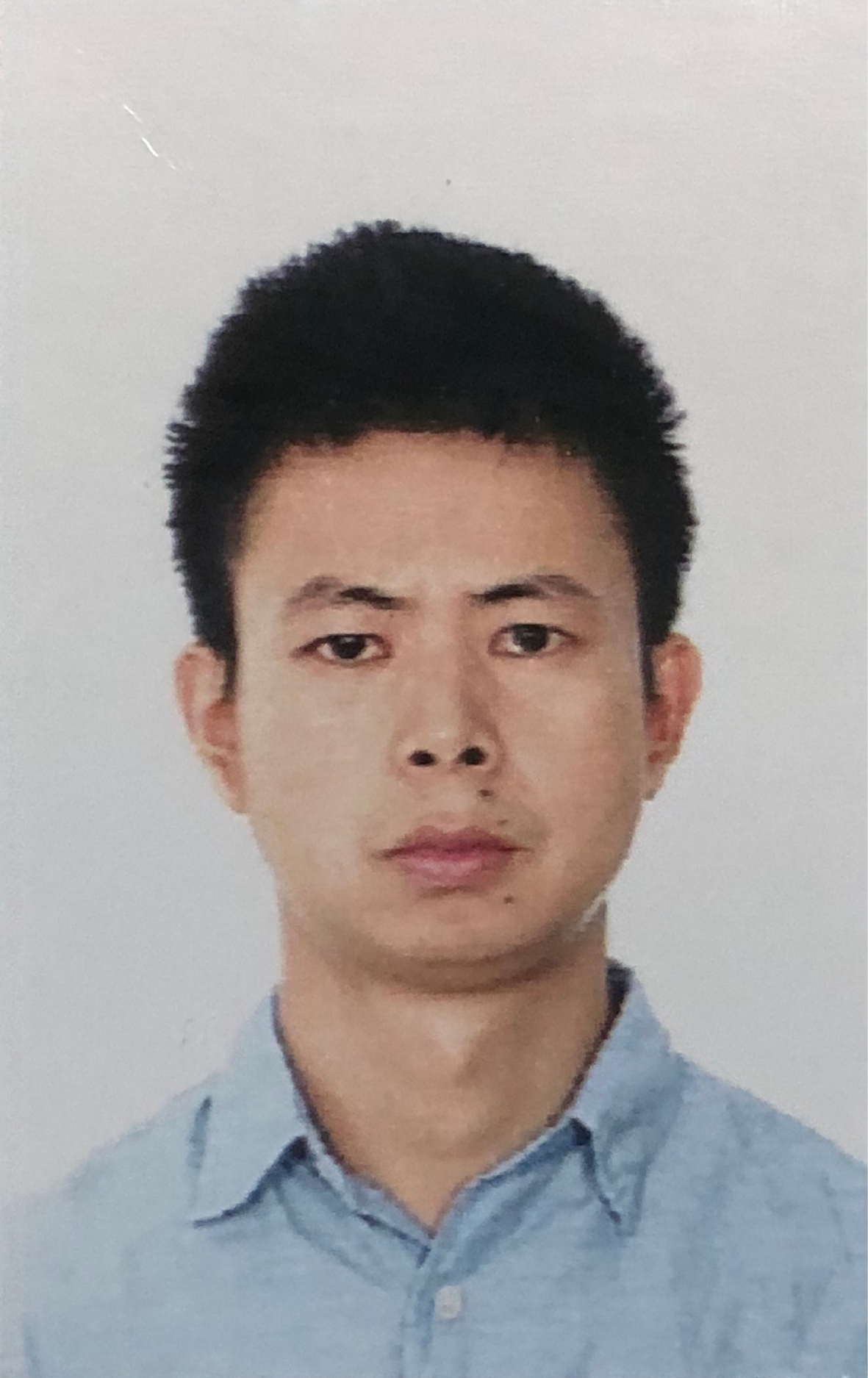}}]
{Xinchu Shi} received the B.E. degree from Huazhong University of Science and Technology (HUST), Wuhan, China, in 2008, and the Ph.D degree in pattern recognition and intelligent system from the Institute of Automation, Chinese Academy of Sciences, Beijing, China, in 2014.

He was a Research Assistant with Temple University, USA, in 2011 and 2013, and worked as an assistant professor in Institute of Automation from 2014 to 2016. He is currently a senior manager in Meituan Autonomous Delivery (MAD). His research interests include machine learning, computer vision and autonomous driving.
\end{IEEEbiography}

\end{document}